\documentclass[preprint,12pt,a4paper]{elsarticle}

\usepackage[a4paper,left=23mm,right=23mm,top=40mm,bottom=32mm]{geometry}
\usepackage{lmodern}
\usepackage{amsmath}
\usepackage{amssymb}
\usepackage{booktabs}
\usepackage{graphicx}
\usepackage{float}
\usepackage{placeins}
\usepackage{subcaption}
\usepackage{multirow}
\usepackage{siunitx}
\usepackage{xurl}
\usepackage{xcolor}
\usepackage{hyperref}
\newcommand{\revone}[1]{#1}
\hypersetup{
    colorlinks=false,
    pdfborder={0 0 1},
    linkbordercolor={1 0 0},
    citebordercolor={0 1 0},
    urlbordercolor={1 1 1},
    filebordercolor={1 1 1}
}

\journal{Theoretical \& Applied Mechanics Letters}

\begin{document}

\begin{frontmatter}

\title{A Geometry-Aware Triplane Field Network for Vehicle Aerodynamic Prediction}

\author[inst1,inst2]{Kangkang Qi}
\author[inst1,inst2]{Huiyu Yang}
\author[inst3]{Keqi Ding\corref{cor2}}
\author[inst1,inst2]{Yunpeng Wang}
\author[inst4]{Yuntian Chen}
\author[inst4,inst3]{Yuanwei Bin}
\author[inst3]{Rikui Zhang}
\author[inst1,inst2]{Jianchun Wang\corref{cor1}}

\cortext[cor1]{Corresponding author: wangjc@sustech.edu.cn}
\cortext[cor2]{Corresponding author: dingkq@tenfong.cn}

\address[inst1]{Department of Mechanics and Aerospace Engineering, Southern University of Science and Technology, Shenzhen 518055, China}
\address[inst2]{Shenzhen Key Laboratory of Complex Aerospace Flows, Southern University of Science and Technology, Shenzhen 518055, China}
\address[inst3]{Shenzhen Tenfong Technology Co., Ltd., Shenzhen 518000, China}
\address[inst4]{Ningbo Key Laboratory of Advanced Manufacturing Simulation, Eastern Institute of Technology, Ningbo 315200, China}

\begin{abstract}
High-fidelity computational fluid dynamics (CFD) is crucial to vehicle aerodynamic analysis, but its cost still constrains early-stage design exploration. Machine-learning-based surface-field prediction offers a faster alternative if the model can efficiently capture both global flow context and local geometric detail. This work proposes a machine-learning-based method, named the geometry-aware triplane field network (GTF-Net), for vehicle aerodynamic pressure and wall shear stress prediction. GTF-Net constructs triplane features directly from sampled surface points through a shared multilayer perceptron (MLP) and smooth bilinear rasterization. The planes are then processed by a dual-stream backbone that combines adaptive Fourier neural operator (AFNO) spectral mixing with convolutional neural network (CNN) refinement, so long-range aerodynamic coupling and local geometry-induced variations are modeled in the same representation. \revone{At the query stage,} sampled triplane features are combined with vehicle-aligned directional coordinates, normal-projection features, and a voxel-based curvature proxy. GTF-Net is compared with Transolver, geometry-informed neural operator (GINO), and TripNet, a triplane-based surrogate model. GTF-Net improves the relative $L_2$ error from the strongest baseline value of 0.157 to 0.145 for pressure prediction and from 0.237 to \revone{0.225} for wall shear stress prediction. Ablation results show that AFNO mixing, local CNN refinement, and query-side geometric encoding each contribute to accuracy, supporting the proposed mechanism of combining structured triplane representation with explicit aerodynamic geometry cues. \revone{Code and data are available at: \url{https://github.com/qkk106329/GTF_Net}.}
\end{abstract}


\end{frontmatter}

\section{Introduction}

High-fidelity computational fluid dynamics (CFD) remains indispensable in vehicle aerodynamic analysis, but its computational cost limits large-scale design iteration. Data-driven surrogates that predict aerodynamic quantities directly from geometry have therefore attracted growing attention. Physics-informed and transfer-learning approaches have been explored for data-scarce fluid and aerodynamic prediction tasks \cite{rao2020pinn,zhang2025transferpressure}. Operator learning methods, including DeepONet and the Fourier neural operator (FNO), learn mappings between function spaces \cite{lu2021deeponet,li2021fno}. More recently, general-geometry neural solvers, including Transolver and the spatially-aware transformer operator (SATO), have demonstrated strong performance on complex partial differential equation (PDE) domains \cite{wu2024transolver,yang2026sato}. Scalable point-cloud and transformer neural operators have further targeted large or variable three-dimensional geometries through million-scale solvers, point-cloud operators, and multiscale CFD architectures \cite{luo2025transolverpp,zeng2025pcno,wang2025mno,zeng2025lrqsolver}.

In automotive aerodynamics, \revone{surrogate modeling has shifted} from simplified canonical bodies toward realistic high-fidelity vehicle geometries based on large-scale datasets. DrivAerNet introduced a parametric car dataset for aerodynamic design and prediction \cite{elrefaie2025drivaernet}, while DrivAerNet++ extended this line with large-scale multimodal CFD simulations and learning benchmarks \cite{elrefaie2024drivaernetpp}. CarBench further broadened the benchmark landscape for neural surrogates in high-fidelity three-dimensional car aerodynamics \cite{elrefaie2025carbench}. Related vehicle-oriented surrogates have progressed from interactive three-dimensional flow prediction to scalable automotive aerodynamic operators \cite{umetani2018flow,alkin2025abupt,choy2025figconv,liu2025aerogto}. Recent studies have further expanded this line of research toward real-world automotive design scenarios, including multi-scale transformer models for drag coefficient estimation, mesh-free CAD-to-flow prediction on non-watertight geometries, and multi-output aerodynamic surrogates for rapid design iteration \cite{liu2025dragsolver,gu2026geoformer,liu2026emos}. Agentic vision--physics--decision frameworks have also begun to connect aerodynamic prediction with design decision making \cite{liu2026aeroagent}. CNN-based studies have also explored real-time aerodynamic evaluation and vehicle flow-field prediction for arbitrary vehicle shapes, cars, and buses \cite{jacob2021realtime,chen2021vehicleflow,garciafernandez2023busflow}.

There are several types of representations for learning on irregular 3D geometry. Point-based architectures, including PointNet, PointNet++, PointNeXt, and Point Transformer, provide flexible encoders for unordered point samples \cite{qi2017pointnet,qi2017pointnetpp,qian2022pointnext,zhao2021pointtransformer}. Graph- and mesh-based models including MeshGraphNets directly exploit discrete geometric connectivity and local interactions \cite{pfaff2021meshgraphnets}. Geometry-aware operator models including geometry-aware Fourier neural operator (Geo-FNO) and GINO further extend neural operators to irregular geometric settings \cite{li2023geofno,li2023gino}. Neural field approaches, exemplified by the Aero-NeF neural field model, represent aerodynamic quantities as continuous implicit functions of spatial coordinates \cite{catalani2024aeronef}. Across these representation families, the core challenge is the same: balancing geometric fidelity, long-range physical context, and computational efficiency on complex surfaces.

Among structured representations, the triplane representation was popularized in 3D-aware generative modeling as a compact way to encode 3D information with three axis-aligned feature planes \cite{chan2022eg3d}. By converting a 3D field into a form amenable to efficient 2D backbones, triplanes offer an appealing compromise between memory efficiency and spatial continuity. In vehicle aerodynamics, it was shown that triplane representations can support drag prediction, surface pressure estimation, and full 3D flow-field inference on DrivAerNet and DrivAerNet++ \cite{chen2025tripnet}.

For the high-fidelity surface-field setting considered here, however, two issues remain open. First, TripNet constructs its triplane representation through an additional object-specific preprocessing and fitting stage before downstream prediction \cite{chen2025tripnet}, whereas a surface-field model can instead seek to build the latent representation directly from sampled surface points. Second, although triplane features support continuous query-based prediction, the role of explicit geometric conditioning at the decoding stage remains underexplored for surface pressure and wall shear stress, whose distributions depend not only on global vehicle shape but also on local orientation and geometric variation.

We address these issues with the geometry-aware triplane field network (GTF-Net), a triplane framework built around three improvements. First, instead of relying on a separate object-specific fitting step, we use a shared multilayer perceptron (MLP) that maps each surface sample to a latent feature before smooth bilinear rasterization onto the three planes. Second, a dual-stream backbone combines adaptive Fourier neural operator (AFNO) spectral mixing \cite{guibas2021afno} with convolutional layers, capturing both long-range aerodynamic correlations and local spatial patterns. Third, a geometry-aware query decoder augments sampled triplane features with vehicle-aligned directional coordinates, normal-projection features, and a local curvature proxy, so explicit geometric conditioning is injected exactly where the field value is predicted.

We evaluate GTF-Net on the DrivAerNet++ surface-field benchmark for both pressure and wall shear stress prediction, comparing against Transolver \cite{wu2024transolver}, GINO \cite{li2023gino}, and TripNet \cite{chen2025tripnet}. The main contributions are:
\begin{itemize}
    \item We propose a triplane field network with a shared MLP and smooth bilinear rasterization that builds triplane features directly from sampled surface points, avoiding a separate object-specific fitting step and providing a richer geometric input to the structured backbone.
    \item We introduce a geometry-aware query decoder that combines sampled triplane features with vehicle-aligned directional encoding, normal-projection features, and a local curvature proxy through learnable residual gating, providing explicit geometric conditioning at the decode stage.
    \item We demonstrate on the DrivAerNet++ surface-field benchmark that the proposed model \revone{improves predictive accuracy relative to} Transolver, GINO, and TripNet on both pressure and wall shear stress prediction.
\end{itemize}

\begin{figure}[!t]
    \centering
    \includegraphics[width=\linewidth]{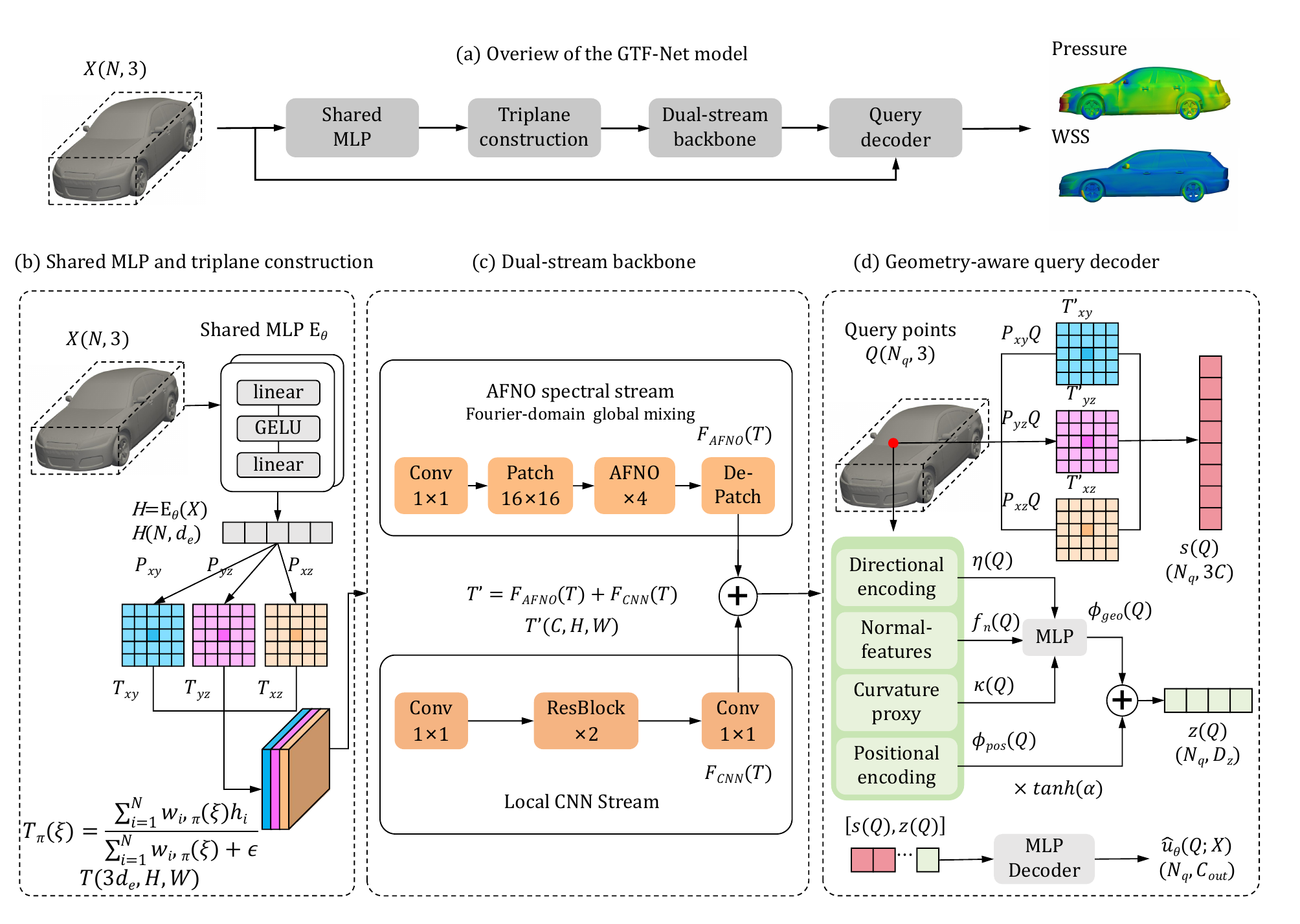}
    \caption{Overview of GTF-Net. Surface samples are mapped to three differentiable triplane feature maps, refined by a dual-stream AFNO--CNN backbone, and decoded at query locations using triplane features together with positional and geometry-aware encodings.}
    \label{fig:method_overview}
\end{figure}

\section{Method}

\subsection{Governing equations and conventional CFD}

In external vehicle aerodynamics, air is commonly modeled as an incompressible fluid governed by the Navier--Stokes equations \cite{pope2000turbulent}:
\begin{equation}
\begin{aligned}
\frac{\partial \boldsymbol{v}}{\partial t}
+ (\boldsymbol{v}\cdot\nabla)\boldsymbol{v}
&= -\revone{\frac{1}{\rho}}\nabla p + \nu \Delta \boldsymbol{v} + \boldsymbol{f},
\qquad \text{in } \Omega, \\
\nabla \cdot \boldsymbol{v}
&= 0,
\qquad \text{in } \Omega .
\end{aligned}
\end{equation}
where $\Omega$ denotes the fluid domain around the vehicle, $\Gamma$ denotes the vehicle surface boundary, $t$ is time, $\boldsymbol{v}$ is the velocity field, $p$ is the \revone{physical pressure}, $\rho$ is the constant fluid density, $\nu$ is the kinematic viscosity, $\nabla$ is the spatial gradient operator, $\Delta$ is the Laplacian operator, and $\boldsymbol{f}$ is an external body force, which is typically neglected in external aerodynamic simulations. \revone{Accordingly, the pressure-gradient term is written as $-(1/\rho)\nabla p$.}

On the vehicle surface $\Gamma$, the standard no-slip condition $\boldsymbol{v}=0$ is imposed, while freestream conditions are prescribed on the far-field boundary.

The flow regime is characterized by the Reynolds number
\begin{equation}
\revone{\mathrm{Re}}=\frac{U_{\infty}L_{c}}{\nu},
\end{equation}
where $U_{\infty}$ is the incoming freestream velocity and $L_{c}$ is a characteristic vehicle length.

For the high-Reynolds-number turbulent flows typical of road vehicles, practical CFD most often solves the Reynolds-averaged Navier--Stokes (RANS) equations.

\begin{equation}
\begin{aligned}
(\bar{\boldsymbol{v}}\cdot\nabla)\bar{\boldsymbol{v}}
&= -\revone{\frac{1}{\rho}}\nabla \bar{p}
+ \nu \Delta \bar{\boldsymbol{v}}
- \nabla \cdot \boldsymbol{\tau}_{R},
\qquad \text{in } \Omega, \\
\nabla \cdot \bar{\boldsymbol{v}}
&= 0,
\qquad \text{in } \Omega, \\
\end{aligned}
\end{equation}
where $\bar{\boldsymbol{v}}$ and $\bar{p}$ are Reynolds-averaged velocity and physical pressure, and $\boldsymbol{\tau}_{R}=\overline{\boldsymbol{v}'\otimes\boldsymbol{v}'}$ is the \revone{kinematic} Reynolds-stress tensor introduced by turbulence averaging. Closing this system requires a turbulence model, including the $k$--$\omega$ SST model commonly used in automotive aerodynamics \cite{menter1994sst}.

\revone{\quad The CFD setup of the DrivAerNet++ dataset is fixed as follows \cite{elrefaie2024drivaernetpp}. All cases use 1:1-scale vehicle geometry, OpenFOAM v11 with the steady incompressible \texttt{simpleFoam} solver, RANS with the $k$--$\omega$ SST turbulence model, air density $\rho=1.184\,\mathrm{kg\,m^{-3}}$, and kinematic viscosity $\nu=1.56\times10^{-5}\,\mathrm{m^2\,s^{-1}}$. The domain specification is also fixed: a velocity inlet with a uniform prescribed velocity of $U_\infty=30\,\mathrm{m\,s^{-1}}$, a pressure outlet with the \texttt{inletOutlet} velocity treatment \cite{openfoam2023inletoutlet}, a no-slip vehicle surface, a moving ground, rotating-wall wheel boundary conditions, slip lateral and top boundaries, and a symmetry plane. Except for the vehicle geometry and category-dependent meshing strategy, the solver and boundary-condition specification are unchanged across the dataset. The Reynolds number based on vehicle length is therefore not a single constant: because the vehicle length varies from 4.344 to 5.210\,m, $\mathrm{Re}$ ranges from $8.366\times10^6$ to $1.006\times10^7$. This geometry-induced variation is part of the design distribution rather than an independently varied operating condition.}

\revone{Consequently, GTF-Net does not take freestream speed, yaw angle, ground speed, wheel-speed ratio, fluid properties, or turbulence-model settings as inputs. Its learned operator is a fixed-condition surrogate; prediction at other flow conditions would require these variables to be supplied as conditioning inputs and the model to be trained on corresponding multi-condition CFD data.}

Classical numerical solvers discretize these equations on computational grids using methods including the finite difference method, finite element method, finite volume method, and spectral method, while alternative approaches including the lattice Boltzmann method solve kinetic equations that recover the same macroscopic behavior \cite{ferziger2002cfd}. In automotive CFD, finite-volume RANS solvers are especially common. They resolve the full volumetric flow around the vehicle, but they remain computationally expensive on high-resolution meshes. This cost motivates machine-learning-based surrogate models that predict target surface quantities directly.

\subsection{Problem formulation}

Our objective is to learn a surrogate model that maps vehicle surface geometry to the corresponding surface field, rather than solving the full flow field in $\Omega$ \cite{li2023gino,catalani2024aeronef,chen2025tripnet}. This mapping can be expressed as:
\begin{equation}
 \revone{\mathcal{F}_{\boldsymbol{c}_0,p_k}:\Gamma \mapsto \bar{p}_k|_{\Gamma}, \qquad \mathcal{F}_{\boldsymbol{c}_0,\tau}:\Gamma \mapsto \boldsymbol{\tau}_w|_{\Gamma}.}
\end{equation}
where $\Gamma$ represents the vehicle surface geometry, \revone{$\boldsymbol{c}_0$ denotes the fixed CFD setting specified above, $p_k=p/\rho$ is the kinematic pressure stored by the incompressible OpenFOAM cases,} and $\boldsymbol{\tau}_w|_{\Gamma}$ is the wall shear stress restricted to that surface. \revone{The physical-pressure convention follows from Eqs.~(1) and (3), whereas $p_k$ denotes the pressure-learning target.}

In practice, the surface is represented by a discrete point sample $X\subset\Gamma$, with $X=\{\boldsymbol{x}_i\}_{i=1}^{N}$, where \(\boldsymbol{x}_i\in\mathbb{R}^{3}\) is a three-dimensional surface coordinate and \(N\) is the number of sampled surface points. We approximate the surface operator with a query-based surrogate
\begin{equation}
\hat{\boldsymbol{u}}_{\theta}(\boldsymbol{x}; X) \approx \boldsymbol{u}(\boldsymbol{x}), \qquad \boldsymbol{x} \in \Gamma.
\end{equation}
Here, \(\boldsymbol{u}\) denotes the generic target surface field, corresponding to \revone{\(\bar{p}_k|_{\Gamma}\)} for pressure prediction and \(\boldsymbol{\tau}_w|_{\Gamma}\) for wall shear stress prediction, while \(\hat{\boldsymbol{u}}_{\theta}\) is the model prediction parameterized by \(\theta\).

During training, the loss is evaluated on a sampled query subset $\boldsymbol{X}_Q \subseteq X$:
\begin{equation}
\mathcal{L}_{\mathrm{MSE}}=\mathrm{MSE}=
\frac{1}{M}
\sum_{i=1}^{M}
\left(y_i-\hat{y}_i\right)^2,
\label{eq:mse_loss}
\end{equation}
where MSE denotes mean squared error, $y_i$ and $\hat{y}_i$ represent the ground-truth and predicted values, respectively, after flattening all evaluated surface points and field components, and \(M\) is the total number of scalar entries.

\subsection{Triplane feature construction}
As shown in Fig.~\ref{fig:method_overview}(b), the input surface geometry is first converted into
a triplane feature tensor. The sampled surface point set \(X=\{\boldsymbol{x}_i\}_{i=1}^{N}\) is used to construct the triplane. In the reported configuration, only
the normalized point coordinates are used as input. Each point
is lifted to a latent feature by a shared multilayer perceptron:
\begin{equation}
\boldsymbol{h}_i = E_{\theta}(\boldsymbol{x}_i), \qquad \boldsymbol{h}_i\in\mathbb{R}^{d_e},
\end{equation}
where \(E_{\theta}\) denotes a shared MLP applied independently to each surface point and \(d_e\) is the point embedding
dimension.

The three-dimensional point \(\boldsymbol{x}_i=(x_i,y_i,z_i)\) is then projected onto
three axis-aligned planes. The projection operators are defined as
\[
P_{xy}(\boldsymbol{x}_i)=(x_i,y_i),\qquad
P_{yz}(\boldsymbol{x}_i)=(y_i,z_i),\qquad
P_{xz}(\boldsymbol{x}_i)=(x_i,z_i).
\]
For a plane \(\pi\in\{xy,yz,xz\}\), let
\(\Lambda_{\pi}\subset\mathbb{R}^{2}\) denote its two-dimensional grid
domain and let \(\boldsymbol{\xi}\in\Lambda_{\pi}\) be a grid location on this plane.
The feature at \(\boldsymbol{\xi}\) is obtained by bilinear weighted rasterization:
\begin{equation}
T_{\pi}(\boldsymbol{\xi})=
\frac{\sum_{i=1}^{N} w_{i,\pi}(\boldsymbol{\xi})\boldsymbol{h}_i}
{\sum_{i=1}^{N} w_{i,\pi}(\boldsymbol{\xi})+\epsilon},
\end{equation}
where \(w_{i,\pi}(\boldsymbol{\xi})\) is the bilinear interpolation weight between the
projected point \(P_{\pi}(\boldsymbol{x}_i)\) and the grid location \(\boldsymbol{\xi}\), and
\(\epsilon=10^{-8}\) is a small constant for numerical stability.

This formulation separates the quantities in three-dimensional space and
those on the two-dimensional planes. Specifically, \(\boldsymbol{x}_i\in\mathbb{R}^{3}\)
is a surface point, \(P_{\pi}(\boldsymbol{x}_i)\in\mathbb{R}^{2}\) and
\(\boldsymbol{\xi}\in\mathbb{R}^{2}\) are plane coordinates, \(\boldsymbol{h}_i\in\mathbb{R}^{d_e}\)
is the learned point feature, and \(T_{\pi}(\boldsymbol{\xi})\in\mathbb{R}^{d_e}\) is
the rasterized triplane feature at that grid location.

After rasterization, unoccupied cells are filled by local average smoothing.
The three planes are concatenated along the channel dimension to form the
triplane tensor
\begin{equation}
T=\mathrm{Concat}(T_{xy},T_{yz},T_{xz})
\in\mathbb{R}^{3d_e\times H\times W}
\end{equation}
where \(H\times W\) is the plane resolution.

\subsection{Dual-stream backbone}
As shown in Fig.~\ref{fig:method_overview}(c), the rasterized triplane tensor is processed by a
dual-stream backbone. The input tensor
\(T\in\mathbb{R}^{3d_e\times H\times W}\) is sent to an AFNO-based
spectral stream and a local CNN refinement stream in parallel. The spectral
stream is used to model long-range aerodynamic correlations on the structured
triplane representation, while the CNN stream refines local spatial patterns
near geometric transitions.

In the spectral stream, the triplane tensor is first lifted to a hidden feature
space through a convolutional patch embedding operator:
\[
Z_0=\mathcal{P}(T)+E_{\mathrm{pos}},
\]
where \(\mathcal{P}\) denotes the patch embedding operator and
\(E_{\mathrm{pos}}\) is a learnable positional embedding on the patch grid.
The hidden feature is then processed by a stack of AFNO blocks following the adaptive Fourier neural operator design \cite{guibas2021afno}. For the
\(\ell\)-th block, the update can be written as
\begin{equation}
\bar{Z}_{\ell}
=
Z_{\ell}
+
\mathcal{A}_{\ell}
\left(
\mathrm{LN}(Z_{\ell})
\right),
\end{equation}
\begin{equation}
Z_{\ell+1}
=
\bar{Z}_{\ell}
+
\mathcal{M}_{\ell}
\left(
\mathrm{LN}(\bar{Z}_{\ell})
\right),
\end{equation}
where \(\mathrm{LN}(\cdot)\) denotes layer normalization,
\(\mathcal{A}_{\ell}\) is the AFNO spectral mixing operator, and
\(\mathcal{M}_{\ell}\) is a pointwise multilayer perceptron. The AFNO
operator performs token mixing in the Fourier domain over the two-dimensional
patch grid:
\begin{equation}
\mathcal{A}_{\ell}(Z)
=
\mathcal{F}^{-1}
\left(
\mathcal{W}_{\ell}
\left(
\mathcal{F}(Z)
\right)
\right),
\end{equation}
where \(\mathcal{F}\) and \(\mathcal{F}^{-1}\) denote the Fourier transform
and inverse Fourier transform, and \(\mathcal{W}_{\ell}\) is the learnable
frequency-domain transformation in the \(\ell\)-th AFNO block. After the
AFNO blocks, the hidden feature is projected back to
the dense triplane resolution:
\begin{equation}
T_{\mathrm{AFNO}}=F_{\mathrm{AFNO}}(T),
\qquad
T_{\mathrm{AFNO}}\in\mathbb{R}^{C\times H\times W}
\end{equation}
where \(C\) is the backbone width.

In parallel, the local CNN stream refines short-range spatial details directly
on the rasterized triplane tensor:
\begin{equation}
T_{\mathrm{CNN}}
=
\mathcal{C}_{\mathrm{out}}
\left(
\mathcal{R}_{B}
\left(
\mathcal{C}_{\mathrm{in}}(T)
\right)
\right),
\qquad
T_{\mathrm{CNN}}\in\mathbb{R}^{C\times H\times W}.
\end{equation}
Here, \(\mathcal{C}_{\mathrm{in}}\) and \(\mathcal{C}_{\mathrm{out}}\) are
\(1\times1\) convolutional projections, and \(\mathcal{R}_{B}\) denotes
\(B\) residual CNN blocks. Each residual block contains two \(3\times3\)
convolutions with group normalization and GELU activation.

The outputs of the two streams are fused by element-wise addition:
\begin{equation}
T' = T_{\mathrm{AFNO}} + T_{\mathrm{CNN}},
\qquad
T'\in\mathbb{R}^{C\times H\times W}.
\end{equation}
The refined feature map \(T'\) is then used for query-based decoding. This
dual-stream design allows the backbone to combine Fourier-domain long-range
mixing with local convolutional refinement.

\subsection{Geometry-aware query decoding}
As shown in Fig.~\ref{fig:method_overview}(d), the decoder predicts the aerodynamic field value
at each query point by combining sampled triplane features with query-side
positional and geometric encodings. Let
\(\boldsymbol{Q}=\{\boldsymbol{q}_j\}_{j=1}^{N_q}\) denote the query surface points, where
\(\boldsymbol{q}_j\in\mathbb{R}^{3}\) and \(N_q\) is the number of query points.

For a query point \(\boldsymbol{q}_j\), the refined feature map \(T'\) is sampled by
bilinear interpolation using the three plane projections:
\begin{equation}
\boldsymbol{s}(\boldsymbol{q}_j)=
\left[
S(T',P_{xy}(\boldsymbol{q}_j)),
S(T',P_{yz}(\boldsymbol{q}_j)),
S(T',P_{xz}(\boldsymbol{q}_j))
\right],
\end{equation}
where \(S(\cdot,\cdot)\) denotes bilinear sampling from the refined feature
map. Since each sampled feature has \(C\) channels, the concatenated triplane
feature satisfies \(\boldsymbol{s}(\boldsymbol{q}_j)\in\mathbb{R}^{3C}\). For all query points,
\(\boldsymbol{s}(\boldsymbol{Q})\in\mathbb{R}^{N_q\times 3C}\).

A Fourier positional encoding is applied to the query coordinate:
\begin{equation}
\boldsymbol{\phi}_{\mathrm{pos}}(\boldsymbol{q}_j)=
\left[
\boldsymbol{q}_j,
\sin(2^0\pi \boldsymbol{q}_j),\cos(2^0\pi \boldsymbol{q}_j),\ldots,
\sin(2^{K_q-1}\pi \boldsymbol{q}_j),\cos(2^{K_q-1}\pi \boldsymbol{q}_j)
\right],
\end{equation}
where \(K_q\) is the number of query Fourier bands, and the sine and cosine functions are applied componentwise to the vector \(\boldsymbol{q}_j\).

The geometric encoding \(\boldsymbol{\phi}_{\mathrm{geo}}(\boldsymbol{q}_j)\) contains directional,
normal-projection, and curvature components. First, a vehicle-aligned
orthonormal frame \((\boldsymbol{d}_{\mathrm{flow}},\boldsymbol{d}_{\mathrm{sym}},\boldsymbol{d}_{\mathrm{vert}})\)
is constructed from the flow direction and the symmetry direction. In the
reported experiments, the default flow and symmetry directions are the
canonical vehicle axes. Let \(\bar{\boldsymbol{q}}_j=\boldsymbol{q}_j-\boldsymbol{c}\), where \(\boldsymbol{c}\) is the
bounding-box center of the query surface points. The directional feature is
defined as
\begin{equation}
\boldsymbol{\eta}(\boldsymbol{q}_j)=
\left[
\frac{\bar{\boldsymbol{q}}_j\cdot \boldsymbol{d}_{\mathrm{flow}}}{\sigma_{\mathrm{flow}}},
\frac{\bar{\boldsymbol{q}}_j\cdot \boldsymbol{d}_{\mathrm{sym}}}{\sigma_{\mathrm{sym}}},
\frac{|\bar{\boldsymbol{q}}_j\cdot \boldsymbol{d}_{\mathrm{sym}}|}{\sigma_{\mathrm{sym}}},
\frac{\bar{\boldsymbol{q}}_j\cdot \boldsymbol{d}_{\mathrm{vert}}}{\sigma_{\mathrm{vert}}}
\right],
\end{equation}
where \(\sigma_{\mathrm{flow}}\), \(\sigma_{\mathrm{sym}}\), and
\(\sigma_{\mathrm{vert}}\) are the corresponding maximum absolute projections
over the query points.

Second, when surface normals are available, normal-projection features are
used in the query-side geometric encoding. Let \(\boldsymbol{n}(\boldsymbol{q}_j)\) denote the unit
normal at \(\boldsymbol{q}_j\). The normal feature is
\begin{equation}
\boldsymbol{f}_n(\boldsymbol{q}_j)=
\left[
\boldsymbol{n}(\boldsymbol{q}_j)\cdot \boldsymbol{d}_{\mathrm{flow}},
\boldsymbol{n}(\boldsymbol{q}_j)\cdot \boldsymbol{d}_{\mathrm{sym}},
\boldsymbol{n}(\boldsymbol{q}_j)\cdot \boldsymbol{d}_{\mathrm{vert}},
|\boldsymbol{n}(\boldsymbol{q}_j)\cdot \boldsymbol{d}_{\mathrm{flow}}|
\right].
\end{equation}
In the reported configuration, surface normals are used in the query decoder
but not in the shared MLP used for triplane construction.

Third, a voxel-based curvature proxy is computed from the query surface
points. The surface is partitioned into a \(G^3\) voxel grid. For the voxel
containing \(\boldsymbol{q}_j\), let \(\mathcal{V}(\boldsymbol{q}_j)\) be the set of surface
points inside the voxel, \(M_j=|\mathcal{V}(\boldsymbol{q}_j)|\), and
\(\bar{\boldsymbol{x}}_j=M_j^{-1}\sum_{\boldsymbol{x}_m\in\mathcal{V}(\boldsymbol{q}_j)}
\boldsymbol{x}_m\). The local covariance matrix is computed as
\begin{equation}
\boldsymbol{\Sigma}(\boldsymbol{q}_j)
=
\frac{1}{M_j}
\sum_{\boldsymbol{x}_m\in\mathcal{V}(\boldsymbol{q}_j)}
(\boldsymbol{x}_m-\bar{\boldsymbol{x}}_j)
(\boldsymbol{x}_m-\bar{\boldsymbol{x}}_j)^{\mathsf{T}}
+10^{-6}\boldsymbol{I},
\end{equation}
where \(\boldsymbol{x}_m\) is a surface point in the voxel and \(\boldsymbol{I}\) is the \(3\times3\) identity matrix. The term \(10^{-6}\boldsymbol{I}\) is added for numerical stability. The curvature proxy is then defined as
\begin{equation}
\kappa(\boldsymbol{q}_j)
=
\frac{\lambda_1}{\lambda_1+\lambda_2+\lambda_3},
\end{equation}
where \(\lambda_1\leq\lambda_2\leq\lambda_3\) are the eigenvalues of the
local covariance matrix. The curvature value is log-normalized before being encoded by a multilayer perceptron.

The directional, normal, and curvature branches are encoded separately and
summed to form the geometric encoding:
\begin{equation}
\boldsymbol{\phi}_{\mathrm{geo}}(\boldsymbol{q}_j)
=
\mathcal{P}_{\mathrm{geo}}
\left(
\boldsymbol{\phi}_{\mathrm{dir}}(\boldsymbol{q}_j)
+
\boldsymbol{\phi}_{\mathrm{normal}}(\boldsymbol{q}_j)
+
\boldsymbol{\phi}_{\mathrm{curv}}(\boldsymbol{q}_j)
\right).
\end{equation}
The positional and geometric encodings are combined through a learnable
residual gate:
\begin{equation}
\boldsymbol{z}(\boldsymbol{q}_j)
=
\boldsymbol{\phi}_{\mathrm{pos}}(\boldsymbol{q}_j)
+
\tanh(\alpha)\boldsymbol{\phi}_{\mathrm{geo}}(\boldsymbol{q}_j),
\end{equation}
where \(\alpha\) is a learnable scalar gate that controls the contribution of the
geometric encoding and is initialized to zero. This initialization allows the
decoder to start from the positional encoding and gradually incorporate
geometric conditioning during training.

The final prediction is produced by a multilayer perceptron:
\begin{equation}
\hat{\boldsymbol{u}}_{\theta}(\boldsymbol{q}_j;X)
=
D_{\theta}
\left(
[\boldsymbol{s}(\boldsymbol{q}_j),\boldsymbol{z}(\boldsymbol{q}_j)]
\right).
\end{equation}
For pressure prediction, \(\hat{\boldsymbol{u}}_{\theta}(\boldsymbol{q}_j;X)\in\mathbb{R}^{1}\), and
for wall shear stress prediction,
\(\hat{\boldsymbol{u}}_{\theta}(\boldsymbol{q}_j;X)\in\mathbb{R}^{3}\).

\subsection{Training objective}
The model is trained with mean squared error on normalized targets. For a sampled query set \(\boldsymbol{X}_Q\), the training objective is
\begin{equation}
\mathcal{L}_{\mathrm{MSE}}
=
\frac{1}{|\boldsymbol{X}_Q|C_u}
\sum_{\boldsymbol{q}_j\in \boldsymbol{X}_Q}
\left\|
\hat{\boldsymbol{u}}_{\theta}(\boldsymbol{q}_j;X)
-
\boldsymbol{u}(\boldsymbol{q}_j)
\right\|_2^2 .
\label{eq:training_objective}
\end{equation}
where \(\boldsymbol{X}_Q\) is the set of sampled query points during training, \(|\boldsymbol{X}_Q|\) is the number of query points, and \(C_u\) is the number of target-field channels. For pressure prediction, \(C_u=1\), and for wall shear stress prediction, \(C_u=3\). The term \(\hat{\boldsymbol{u}}_{\theta}(\boldsymbol{q}_j;X)\) is the model prediction at query point \(\boldsymbol{q}_j\) conditioned on the surface sample \(X\), and \(\boldsymbol{u}(\boldsymbol{q}_j)\) is the corresponding CFD ground truth.

The normalization statistics, subsampling strategy, and other training details are described in Section~3.

\section{Experimental setup}

\subsection{Dataset and task}
All numerical experiments are conducted on the DrivAerNet++ surface-field benchmark \cite{elrefaie2024drivaernetpp}. DrivAerNet++ is a large-scale multimodal automotive aerodynamics dataset generated from high-fidelity CFD simulations of diverse vehicle configurations, including fastback, notchback, and estateback bodies with variations in underbody and wheel design. In addition to aerodynamic coefficients, the dataset provides geometric representations and CFD-derived flow and surface fields; this work focuses on the surface fields associated with pressure and wall shear stress. We evaluate two tasks: surface pressure prediction and wall shear stress (WSS) prediction. The dataset is split into training, validation, and test sets at approximately 71.7\%, 14.1\%, and 14.2\%, corresponding to 5{,}819, 1{,}148, and 1{,}154 samples, respectively.

\subsection{Preprocessing}
The preprocessing includes geometry normalization, target standardization, surface-normal estimation, and curvature-proxy construction. Surface coordinates are normalized to the range $[-1,1]$ \cite{pedregosa2011scikit} as
\begin{equation}
\boldsymbol{x}^{\mathrm{norm}}=
2\frac{\boldsymbol{x}-\boldsymbol{b}_{\min}}{\boldsymbol{b}_{\max}-\boldsymbol{b}_{\min}}-1.
\end{equation}
where $\boldsymbol{x}^{\mathrm{norm}}$ is the normalized coordinate, $\boldsymbol{x}$ is the original coordinate, and $\boldsymbol{b}_{\min}$ and $\boldsymbol{b}_{\max}$ are the training-set componentwise coordinate bounds.
Target fields are standardized by z-score normalization \cite{pedregosa2011scikit},
\begin{equation}
\tilde{\boldsymbol{u}}=\frac{\boldsymbol{u}-\boldsymbol{\mu}_u}{\boldsymbol{\sigma}_u+\epsilon},
\end{equation}
where $\tilde{\boldsymbol{u}}$ is the standardized target field, $\boldsymbol{u}$ is the original target field, $\boldsymbol{\mu}_u$ and $\boldsymbol{\sigma}_u$ are the training-set mean and standard deviation, and \(\epsilon=10^{-8}\) avoids division by very small values.

Point normals are computed from the mesh with automatic orientation. For a surface point $\boldsymbol{x}_i$, the normal is obtained by aggregating the normals of its incident faces,
\begin{equation}
\boldsymbol{n}_i=
\frac{\sum_{f\in\mathcal{N}(i)} A_f\,\boldsymbol{n}_f}
{\left\|\sum_{f\in\mathcal{N}(i)} A_f\,\boldsymbol{n}_f\right\|_2},
\end{equation}
where $\mathcal{N}(i)$ is the set of faces incident to point $\boldsymbol{x}_i$, $\boldsymbol{n}_f$ is the unit normal of face $f$, and $A_f$ is its area.

The local curvature proxy is precomputed on a $16^3$ voxel grid as described in Section~2. During training, each sample is randomly subsampled to 20{,}000 surface points; validation and testing use the full-resolution surface.

\subsection{Baselines}
We compare against Transolver \cite{wu2024transolver}, GINO \cite{li2023gino}, and TripNet \cite{chen2025tripnet}. Transolver is a general-geometry neural PDE solver that uses physics-aware attention with learnable slicing to handle irregular point sets. GINO extends neural operator learning to general geometric domains. TripNet is the closest structured-representation baseline because it also uses triplane features and query-based field decoding. In the original TripNet pipeline, however, each vehicle first undergoes a separate object-specific occupancy-fitting stage to obtain its triplanes, which are then processed by a U-Net-style backbone and queried with an MLP for surface-field prediction. In contrast, GTF-Net constructs triplane features directly from sampled surface points through a shared MLP and smooth bilinear rasterization, without an additional fitting stage, and further adds dual-stream AFNO--CNN processing and query-side geometric conditioning. For comparison under the shared DrivAerNet++ surface-field protocol, we align the data splits, target definitions, normalization pipeline, evaluation metrics, and available training or inference protocols across methods. Each baseline retains its original model design; only experiment-level settings required by the common protocol are adjusted. \revone{The selected baselines represent point-based physics-aware attention, geometry-informed neural operators, and triplane-based query decoding. Diffusion models, ConvNeXt backbones, and POD-based reduced-order models are not included because no publicly available, task-matched implementations provide the same DrivAerNet++ surface-pressure and WSS prediction protocol; each would require additional method-specific representation or training choices.}

\revone{Under the configurations used in this study, the trainable parameter counts are approximately 7.61M for GTF-Net, 2.47M for Transolver, 134.50M for GINO, and 24.10M for TripNet; the TripNet count excludes its separate object-specific triplane-fitting stage.}

\subsection{Implementation details}
The key architectural and training hyperparameters are summarized in Table~\ref{tab:config}. The point-to-triplane construction module uses three \(128\times128\) planes with a 16-dimensional point embedding, while the backbone combines a four-layer AFNO stream of width 128 with a two-block local CNN stream of width 64. The query decoder uses 10 Fourier bands for positional encoding, 4 Fourier bands for geometric encoding, and a four-layer MLP of width 128. During training, 20{,}000 surface points are sampled per vehicle, and the model is optimized for 200 epochs using AdamW with a learning rate of \(10^{-3}\) and weight decay of \(10^{-5}\). Unless otherwise noted, the pressure and wall shear stress models use the same architecture and training schedule, differing only in the output dimension of the final prediction head. All numerical experiments are conducted on a single NVIDIA A100 GPU with 40\,GB memory.

\begin{table}[htbp]
    \centering
    \small
    \setlength{\tabcolsep}{4.5pt}
    \caption{Key architectural and training hyperparameters used in the numerical experiments.}
    \label{tab:config}
    \begin{tabular}{ll}
        \toprule
        Hyperparameter & Value \\
        \midrule
        \multicolumn{2}{l}{\textbf{Triplane encoder}} \\
        Plane resolution ($H\times W$) & $128 \times 128$ \\
        Point MLP width / depth & $32$ / $4$ \\
        Point embedding dimension ($d_e$) & $16$ \\
        Empty-cell smoothing kernel & $5 \times 5$ \\
        \midrule
        \multicolumn{2}{l}{\textbf{Backbone}} \\
        AFNO width / layers & $128$ / $4$ \\
        AFNO patch size / padding & $16$ / $6$ \\
        AFNO channel blocks & $8$ \\
        Local CNN width / residual blocks ($B$) & $64$ / $2$ \\
        \midrule
        \multicolumn{2}{l}{\textbf{Query encoding}} \\
        Positional Fourier bands ($K_q$) & $10$ \\
        Geometric Fourier bands & $4$ \\
        Curvature voxel grid ($G^3$) & $16^3$ \\
        \midrule
        \multicolumn{2}{l}{\textbf{Decoder}} \\
        MLP width / layers & $128$ / $4$ \\
        \midrule
        \multicolumn{2}{l}{\textbf{Training}} \\
        Optimizer / learning rate / weight decay & AdamW / $10^{-3}$ / $10^{-5}$ \\
        Training points per sample ($|\boldsymbol{X}_Q|$) & $20{,}000$ \\
        Batch size / accumulation / epochs & $16$ / $4$ / $200$ \\
        \bottomrule
    \end{tabular}
\end{table}

\subsection{Evaluation metrics}
We adopt relative $L_2$ error and the coefficient of determination ($R^2$) as the main dimensionless evaluation metrics. Metrics are computed on denormalized predictions after flattening all evaluated surface points and field components. Let $\boldsymbol{y}\in\mathbb{R}^{N}$ and $\hat{\boldsymbol{y}}\in\mathbb{R}^{N}$ denote the flattened ground-truth and predicted arrays, respectively, where $N$ is the total number of scalar entries.

The coefficient of determination measures how much of the ground-truth variance is explained by the prediction:
\begin{equation}
R^2 = 1-\frac{\sum_{i=1}^{N}(y_i-\hat{y}_i)^2}{\sum_{i=1}^{N}(y_i-\bar{y})^2},
\end{equation}
where $\bar{y}$ denotes the mean of the ground-truth values.

The relative $L_2$ error measures the normalized Euclidean difference between predictions and ground truth:
\begin{equation}
\mathrm{Relative}\ L_2\ \mathrm{Error} = \frac{\|\hat{\boldsymbol{y}}-\boldsymbol{y}\|_2}{\|\boldsymbol{y}\|_2}.
\end{equation}
For wall shear stress, the three vector components are included in the flattened array before each metric is computed, so the reported values measure joint accuracy across all components.
\section{Results and discussion}

\subsection{Surface pressure prediction}
We first consider the high-fidelity surface pressure task. Table~\ref{tab:main_results} summarizes the quantitative comparison of GTF-Net with Transolver, GINO, and TripNet. For pressure prediction, GTF-Net gives the lowest relative $L_2$ error and highest $R^2$, reducing relative $L_2$ from the strongest baseline value of 0.157 (GINO) to 0.145.

\begin{figure}[!t]
    \centering
    \includegraphics[width=\linewidth]{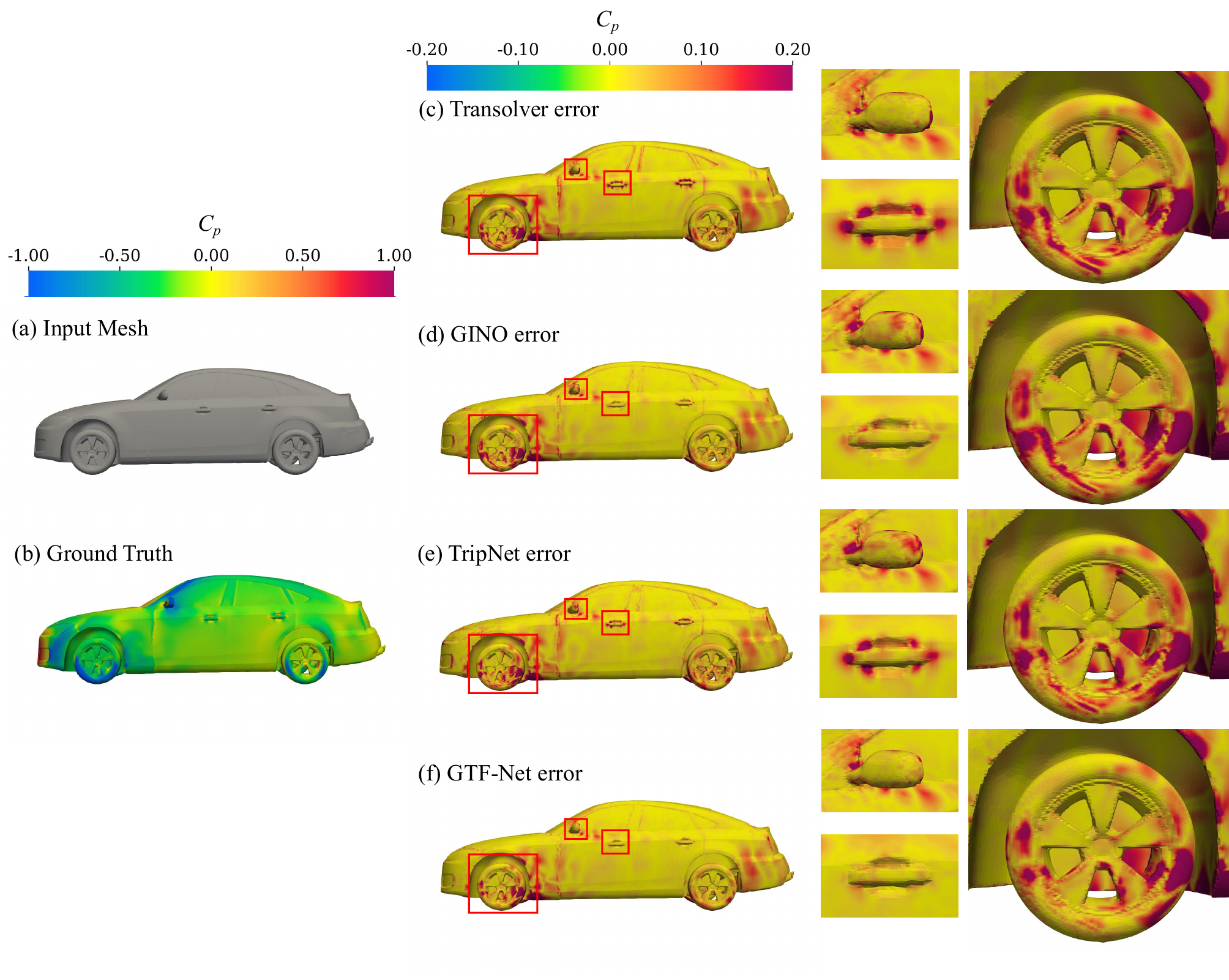}
    \caption{Visualization of the performance of different models in predicting the pressure coefficient \(C_p\) on a DrivAerNet++ test sample: (a) input mesh; (b) ground-truth \(C_p\); (c) error of Transolver; (d) error of GINO; (e) error of TripNet; (f) error of GTF-Net.}
    \label{fig:pressure_cloud_compare}
\end{figure}

Fig.~\ref{fig:pressure_cloud_compare} presents a qualitative comparison of different models on a representative test sample, with pressure visualized as the nondimensional coefficient \(C_p\). The largest pressure errors are concentrated near sharp geometric transitions, including the mirror, door handle, and wheel-arch region. Compared with the baselines, GTF-Net produces smaller and more localized errors in these high-gradient areas, consistent with the aggregate results in Table~\ref{tab:main_results}.

For this visualization, the pressure field is expressed as the nondimensional pressure coefficient \cite{elrefaie2024drivaernetpp},
\begin{equation}
C_p = \revone{\frac{p_k-p_{k,\infty}}{\frac{1}{2}U_{\infty}^{2}}
= \frac{p-p_{\infty}}{\frac{1}{2}\rho U_{\infty}^{2}}},
\end{equation}
where \revone{\(p_k=p/\rho\) and \(p_{k,\infty}=p_{\infty}/\rho\) are the kinematic pressures stored in the incompressible OpenFOAM data, while \(p_{\infty}\), \(\rho\), and \(U_{\infty}\) denote physical freestream pressure, air density, and freestream velocity, respectively.} The error maps show the prediction error between the predicted and CFD-reference \(C_p\) fields.

\begin{table}[htbp]
    \centering
    \small
    \setlength{\tabcolsep}{4.5pt}
    \caption{Dimensionless quantitative comparison on the DrivAerNet++ test split for surface pressure and wall shear stress prediction. Lower is better for relative $L_2$; higher is better for $R^2$. \revone{The metrics are computed separately for each test vehicle and then averaged over the 1,154 vehicles. All methods use 10 independent training runs with random seeds 42--51.}}
    \label{tab:main_results}
    \begin{tabular}{llcc}
        \toprule
        Task & Method & Rel.\ $L_2$ & $R^2$ \\
        \midrule
        \multirow{4}{*}{Pressure} & GINO & \revone{$0.1578 \pm 0.0006$} & \revone{$0.9580 \pm 0.0003$} \\
         & Transolver & \revone{$0.1617 \pm 0.0005$} & \revone{$0.9559 \pm 0.0003$} \\
         & TripNet & \revone{$0.1734 \pm 0.0003$} & \revone{$0.9526 \pm 0.0001$} \\
         & \textbf{GTF-Net} & \textbf{\revone{$0.1445 \pm 0.0004$}} & \textbf{\revone{$0.9644 \pm 0.0002$}} \\
        \midrule
        \multirow{4}{*}{WSS} & Transolver & \revone{$0.2350 \pm 0.0003$} & \revone{$0.9336 \pm 0.0002$} \\
        & GINO & \revone{$0.2443 \pm 0.0010$} & \revone{$0.9277 \pm 0.0006$} \\
         & TripNet & \revone{$0.2583 \pm 0.0003$} & \revone{$0.9272 \pm 0.0001$} \\
         & \textbf{GTF-Net} & \textbf{\revone{$0.2254 \pm 0.0003$}} & \textbf{\revone{$0.9432 \pm 0.0001$}} \\
        \bottomrule
    \end{tabular}
\end{table}

\begin{figure}[!t]
    \centering
        \includegraphics[width=\linewidth]{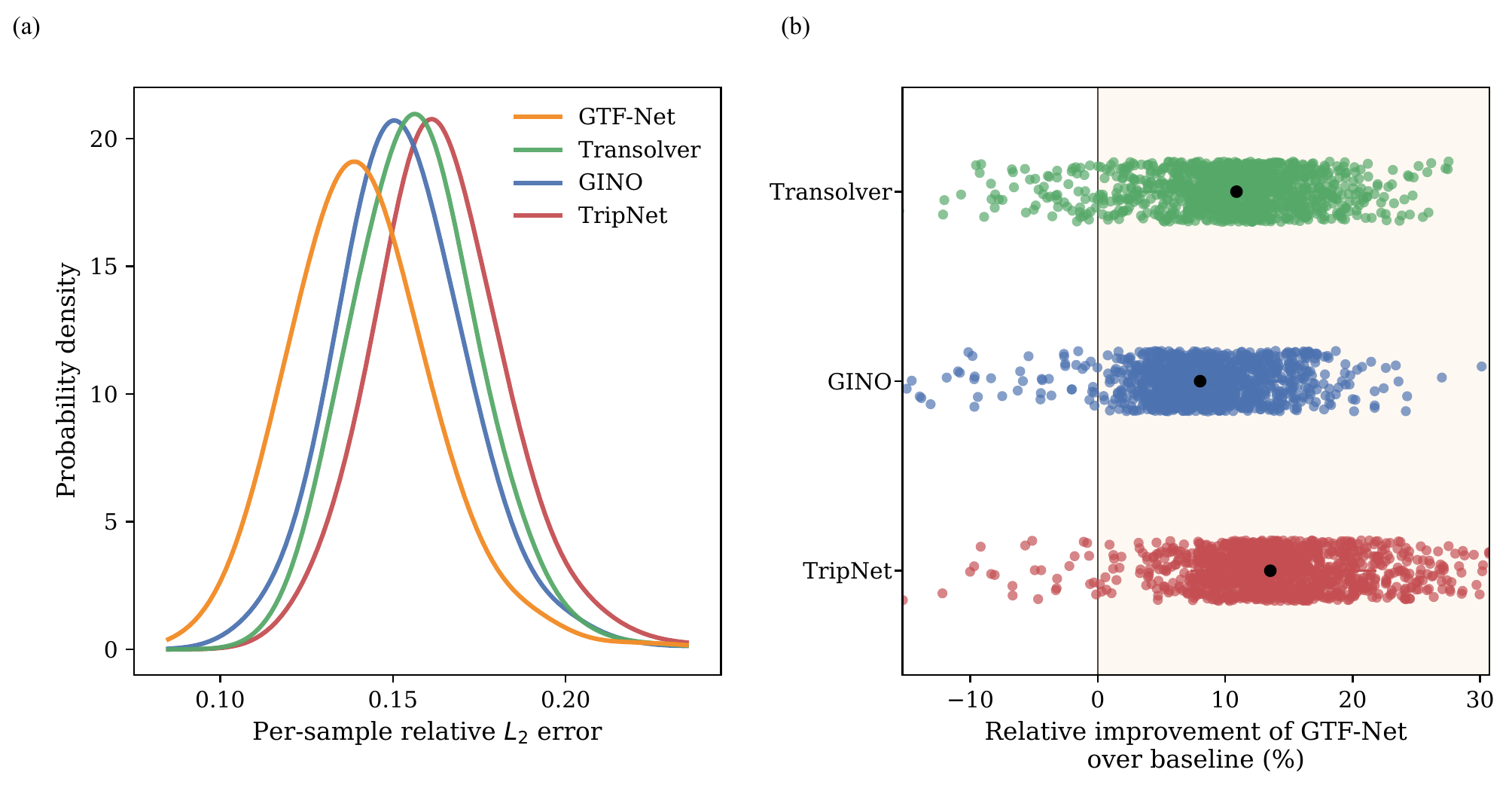}
    \caption{Per-sample surface-pressure prediction performance on the DrivAerNet++ test set. (a) Probability density of relative $L_2$ error for GTF-Net, Transolver, GINO, and TripNet. (b) Relative improvement of GTF-Net over each baseline for individual test samples; colored dots denote individual test samples, black dots denote the median improvement for each baseline, and positive values denote cases where GTF-Net is more accurate.}
    \label{fig:pressure_distribution}
\end{figure}

To check whether this improvement is consistent across designs, Fig.~\ref{fig:pressure_distribution} compares the per-sample relative $L_2$ errors of GTF-Net and the baselines on the full test set. The probability density function (PDF) in Fig.~\ref{fig:pressure_distribution}(a) shows that GTF-Net has lower errors across most test samples, rather than only on a small number of outlier cases. Fig.~\ref{fig:pressure_distribution}(b) reports the per-sample relative improvement of GTF-Net over each baseline,
\begin{equation}
I_m = \frac{e_m-e_{\mathrm{GTF}}}{e_m},
\end{equation}
where $e_m$ is the per-sample relative $L_2$ error of baseline $m$ and $e_{\mathrm{GTF}}$ is the corresponding error of GTF-Net. Positive values therefore denote test samples for which GTF-Net is more accurate.

\FloatBarrier

\begin{figure}[!t]
    \centering
    \includegraphics[width=\linewidth]{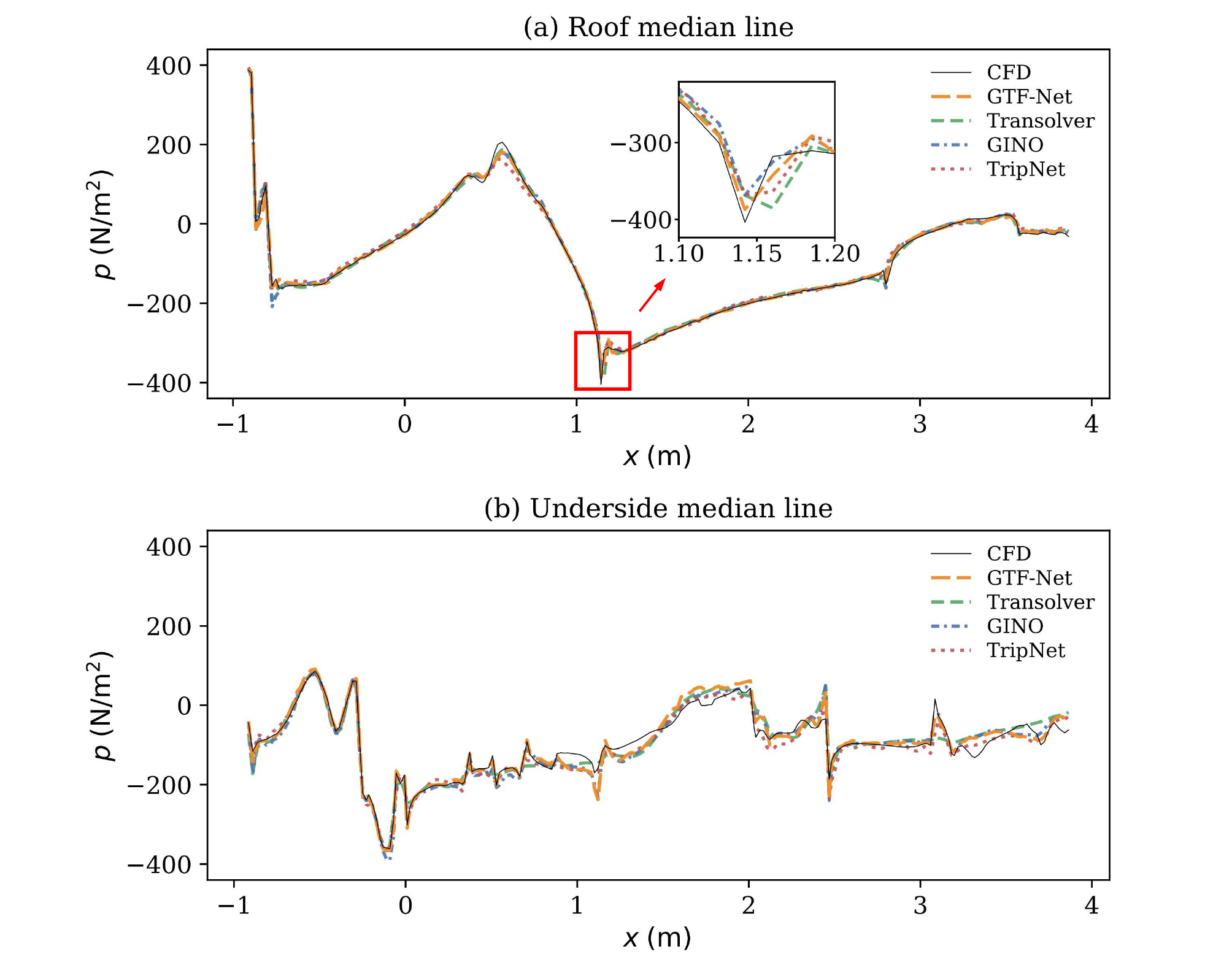}
    \caption{Comparison of the predicted surface pressure along the median line on the roof (a) and underside (b) of a DrivAerNet++ test sample for different models.}
    \label{fig:pressure_line_profiles}
\end{figure}

Fig.~\ref{fig:pressure_line_profiles} shows pressure profiles along the roof and underside median lines of the same sample; GTF-Net remains closer to the CFD reference near sharp pressure changes and recovery regions.

\FloatBarrier

\subsection{Wall shear stress prediction}
Wall shear stress provides a complementary test because it is a vector-valued field and is more sensitive to local surface orientation. Table~\ref{tab:main_results} shows that GTF-Net gives the lowest WSS errors among the compared methods, reducing relative $L_2$ from 0.237 for Transolver, 0.244 for GINO, and 0.258 for TripNet to \revone{0.225}. Taken together with the pressure results, this observation suggests that the triplane pipeline with geometry-aware decoding transfers effectively across both scalar and vector-valued surface-field tasks in the present evaluation setting.

\revone{The WSS decoder produces an unconstrained three-component vector and therefore does not impose a hard tangentiality constraint. We therefore add a tangent-plane diagnostic for GTF-Net on all 1{,}154 test vehicles. For a WSS vector $\boldsymbol{\tau}$ and reconstructed local unit mesh normal $\hat{\boldsymbol{n}}$, the normal component is defined by}
\begin{equation}
\revone{\boldsymbol{\tau}_{n}=(\boldsymbol{\tau}\cdot\hat{\boldsymbol{n}})\hat{\boldsymbol{n}}}
\end{equation}
\revone{The normal leakage is given by}
\begin{equation}
\revone{L_n=100\frac{\|\boldsymbol{\tau}_{n}\|_2}{\|\boldsymbol{\tau}\|_2+\epsilon}}
\end{equation}
\revone{where the norms aggregate all surface points of one vehicle and $\epsilon=10^{-12}$ prevents division by zero. The predicted and reference leakages are $14.926\%\pm7.463\%$ and $15.094\%\pm7.548\%$, respectively, reported as mean $\pm$ standard deviation over vehicles. The VTK files do not contain normals, so the diagnostic uses mesh normals reconstructed during preprocessing. It consequently assesses agreement relative to these discrete normals, rather than exact tangentiality of the continuous CFD field. No post-hoc tangent-plane projection is applied to Table~\ref{tab:main_results}.}

\begin{figure}[!t]
    \centering
    \includegraphics[width=\linewidth]{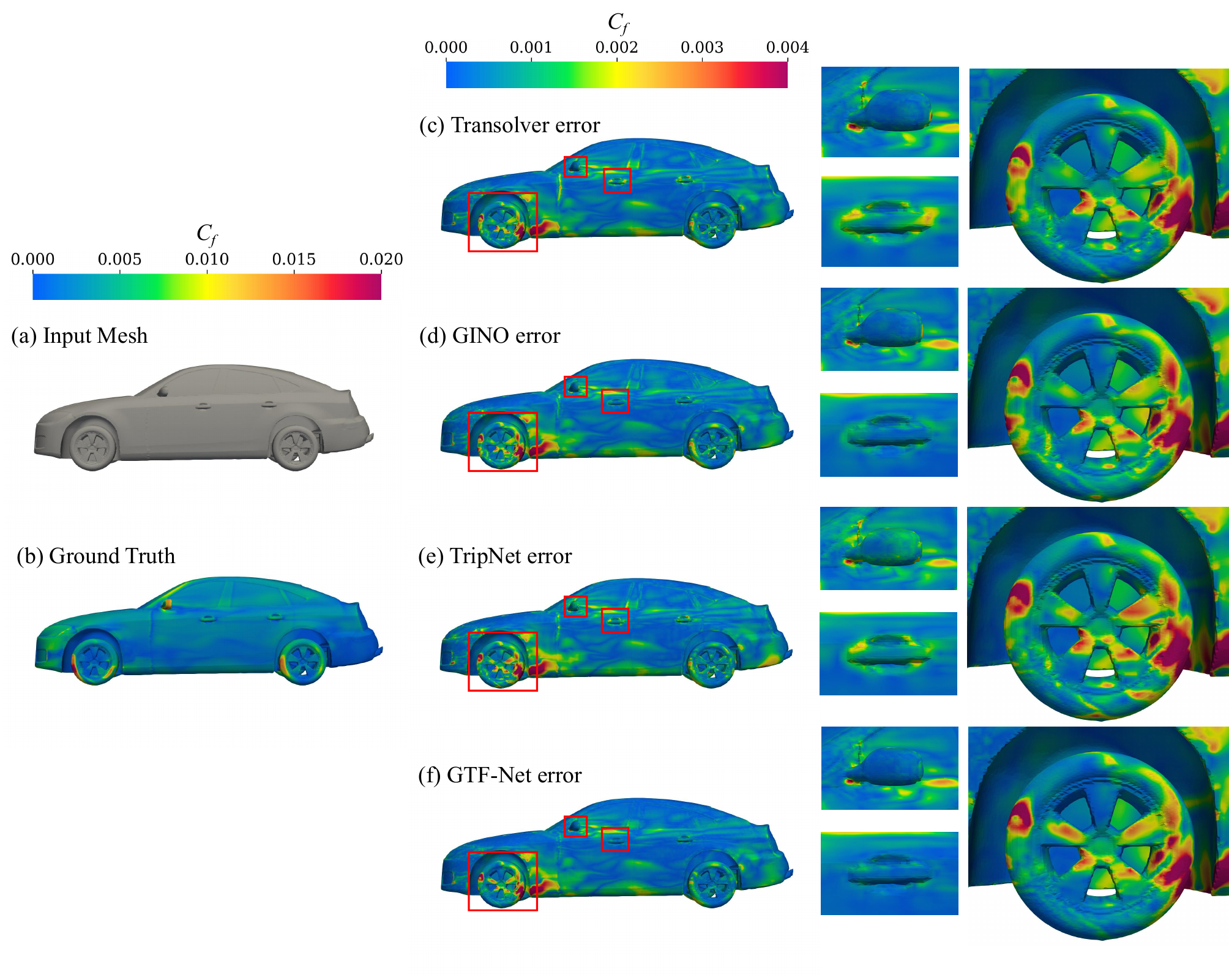}
    \caption{Visualization of the performance of different models in predicting the wall-friction coefficient \(C_f\) on a DrivAerNet++ test sample: (a) input mesh; (b) ground-truth \(C_f\); (c) error of Transolver; (d) error of GINO; (e) error of TripNet; (f) error of GTF-Net.}
    \label{fig:wss_cloud_compare}
\end{figure}

Fig.~\ref{fig:wss_cloud_compare} provides a qualitative comparison for WSS using the same layout and model set as the pressure visualization, with WSS magnitude visualized as the nondimensional coefficient \(C_f\). Because WSS is a vector quantity with stronger local variation, the error magnitudes are generally larger than those for pressure. Prediction errors tend to appear in regions with complex three-dimensional separation, including the underbody, wheel region, and rear diffuser, where local surface orientation changes rapidly. Compared with the baselines, GTF-Net produces smaller and more localized WSS errors in these geometrically complex regions.

For this visualization, the wall shear stress magnitude is expressed as the wall-friction coefficient \cite{elrefaie2024drivaernetpp},
\begin{equation}
C_f = \frac{|\boldsymbol{\tau}_w|}{\frac{1}{2}\rho U_{\infty}^{2}}.
\end{equation}
where \(\boldsymbol{\tau}_w\) is the wall shear stress vector, and \(\rho\) and \(U_{\infty}\) are the air density and freestream velocity, respectively. The error maps show the prediction error between the predicted and CFD-reference \(C_f\) fields.

\FloatBarrier

\begin{figure}[!t]
    \centering
        \includegraphics[width=\linewidth]{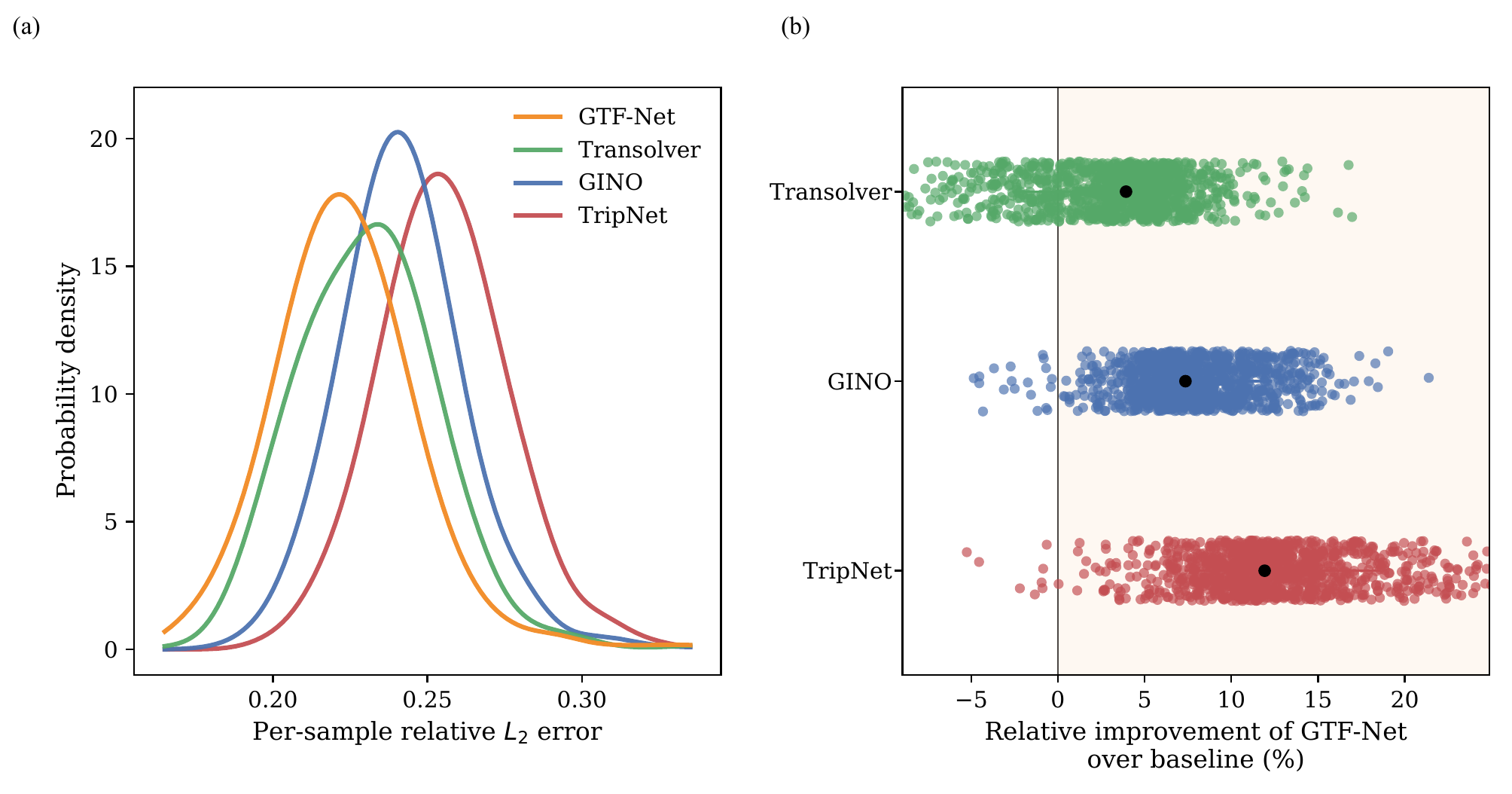}
    \caption{Per-sample wall shear stress prediction performance on the DrivAerNet++ test set. (a) Probability density of relative $L_2$ error for GTF-Net, Transolver, GINO, and TripNet. (b) Relative improvement of GTF-Net over each baseline for individual test samples; colored dots denote individual test samples, black dots denote the median improvement for each baseline, and positive values denote cases where GTF-Net is more accurate.}
    \label{fig:wss_distribution}
\end{figure}

Fig.~\ref{fig:wss_distribution} further compares the per-sample relative $L_2$ errors for WSS prediction. The probability density function (PDF) in Fig.~\ref{fig:wss_distribution}(a) shows the error distribution over the full test set, while Fig.~\ref{fig:wss_distribution}(b) reports the per-sample relative improvement of GTF-Net over each baseline using the same definition as in the pressure analysis. GTF-Net gives the most favorable distribution among the compared models, with positive median improvements over all baselines.

\FloatBarrier

\begin{figure}[!t]
    \centering
    \includegraphics[width=\linewidth]{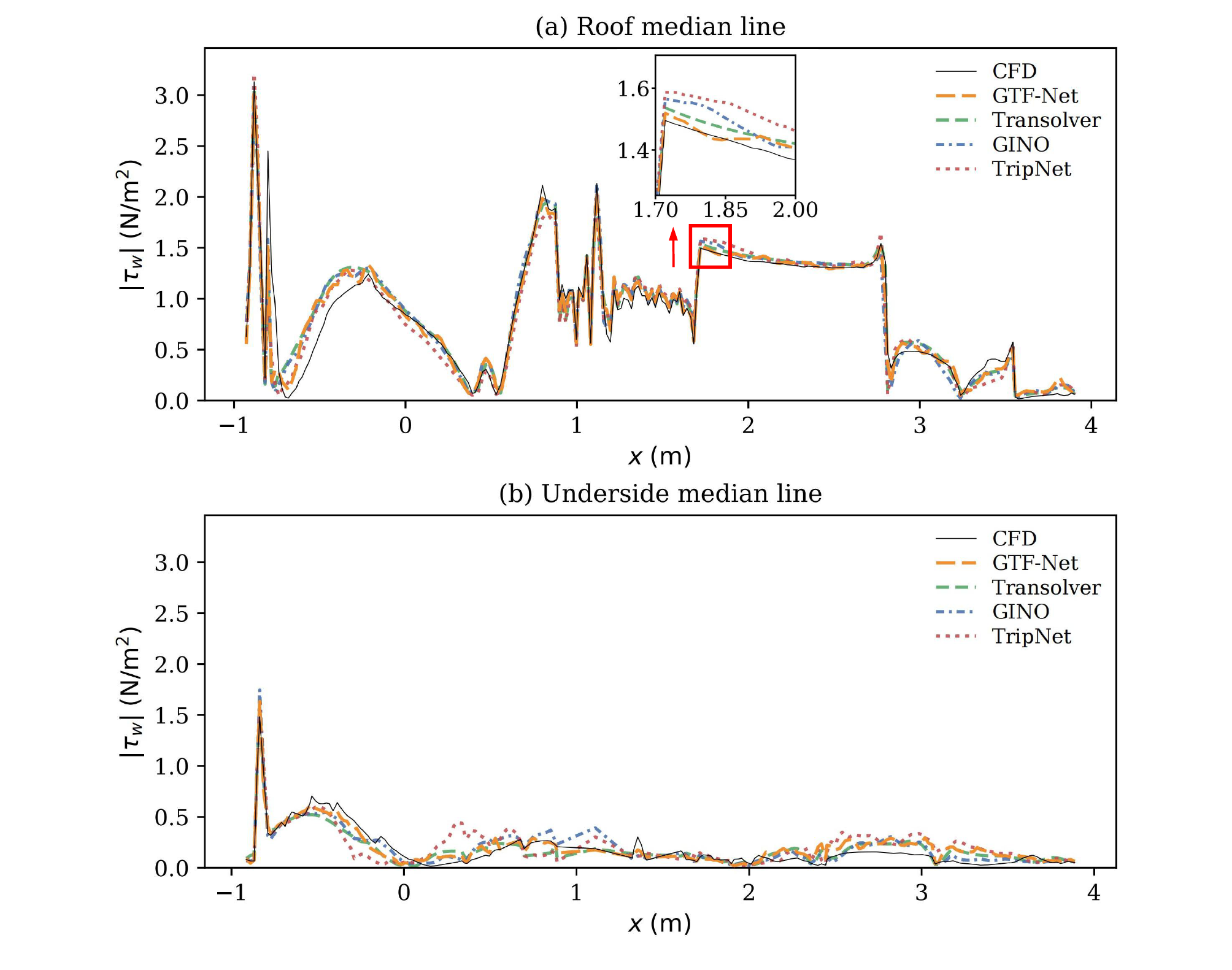}
    \caption{Comparison of the predicted wall shear stress magnitude along the median line on the roof (a) and underside (b) of a DrivAerNet++ test sample for different models.}
    \label{fig:wss_line_profiles}
\end{figure}

Fig.~\ref{fig:wss_line_profiles} shows WSS magnitude profiles along the roof and underside median lines of the same sample. GTF-Net follows the CFD reference more closely than the baselines along both profiles, especially near regions with rapid WSS variation.

\FloatBarrier
\subsection{Ablation study}
We next analyze model components on the prediction of pressure, where the complete ablation suite is run and where localized surface discontinuities make architectural differences particularly visible. Table~\ref{tab:ablation} shows that both streams of the backbone are important. Removing the AFNO branch increases the relative $L_2$ error from 0.145 to 0.157, indicating that local convolutional refinement benefits from global spectral mixing. Removing the CNN branch causes a larger degradation, with relative $L_2$ error increasing to 0.198, showing that local refinement is crucial for resolving pressure variations near geometric transitions.

\revone{Table~\ref{tab:ablation} also isolates the direct point-to-triplane construction stage. Without the shared MLP, the relative $L_2$ error increases to 0.192 and $R^2$ decreases to 0.939. Replacing bilinear rasterization with nearest rasterization and removing empty-cell smoothing both increase the relative $L_2$ error to approximately 0.177. These results support the contribution of learned point features, smooth rasterization, and empty-cell smoothing to the triplane representation.}

The geometric query encoding also contributes. Removing the curvature proxy alone is more harmful than removing the entire geometric encoding branch. To examine this asymmetry, we also remove both normal projections and the curvature proxy while retaining only directional encoding. This variant recovers to the level of the full geometric-encoding ablation, suggesting that the large effect of the curvature proxy is tied to how normal-projection features are used. Finally, the reduced backbone/query configuration degrades performance markedly, indicating that the default capacity is appropriate for the DrivAerNet++ pressure task.

\begin{table}[htbp]
    \centering
    \small
    \setlength{\tabcolsep}{4.5pt}
    \caption{\revone{Ablation study on the DrivAerNet++ pressure task, including the direct point-to-triplane construction stage.} Each row removes or modifies one design component relative to the full model.}
    \label{tab:ablation}
    \begin{tabular}{lcc}
        \toprule
        Variant & Rel.\ $L_2$ & $R^2$ \\
        \midrule
        \textbf{Full model} & \textbf{0.145} & \textbf{0.964} \\
        \revone{Without shared MLP} & \revone{0.192} & \revone{0.939} \\
        \revone{Without bilinear rasterization} & \revone{0.177} & \revone{0.948} \\
        \revone{Without empty-cell smoothing} & \revone{0.177} & \revone{0.948} \\
        Without CNN branch & 0.198 & 0.938 \\
        Without AFNO branch & 0.157 & 0.959 \\
        Without geometric encoding & 0.147 & 0.963 \\
        Without normal projection & 0.148 & 0.963 \\
        Without curvature proxy & 0.172 & 0.951 \\
        Without normal + curvature & 0.148 & 0.963 \\
        Reduced backbone/query configuration & 0.170 & 0.952 \\
        \bottomrule
    \end{tabular}
\end{table}

\revone{To complement the pressure ablations, we next analyze model components on the prediction of WSS, where the vector-valued target provides a complementary test of the proposed architecture. Table~\ref{tab:wss_ablation} shows that both streams of the backbone are important. Removing the AFNO branch increases the relative $L_2$ error from 0.226 to 0.250, indicating that global spectral mixing is important for modeling long-range WSS correlations. Removing the CNN branch causes a smaller degradation, with relative $L_2$ error increasing to 0.227, showing that local refinement still contributes to the vector-field prediction.}

\revone{Table~\ref{tab:wss_ablation} also isolates the direct point-to-triplane construction stage. Without the shared MLP, the relative $L_2$ error increases to 0.261 and $R^2$ decreases to 0.924. Replacing bilinear rasterization with nearest rasterization gives relative $L_2=0.252$ and $R^2=0.929$, and removing empty-cell smoothing gives relative $L_2=0.252$ and $R^2=0.929$. These results support the contribution of learned point features, smooth rasterization, and empty-cell smoothing to the triplane representation for WSS prediction.}

\revone{The geometric query encoding also contributes to the field prediction. Removing either the physical positional encoding or the normal-projection features increases the relative $L_2$ error to 0.227. To examine their joint contribution, we also remove both normal projections and the curvature proxy while retaining only directional encoding. This variant gives relative $L_2=0.227$ and $R^2=0.942$, suggesting that the physical encoding components provide a small but consistent benefit to WSS prediction. The curvature proxy alone has no measurable effect in this configuration. Finally, the reduced backbone/query configuration degrades performance markedly, indicating that the default capacity is appropriate for the DrivAerNet++ WSS task.}

\begin{table}[htbp]
    \centering
    \small
    \setlength{\tabcolsep}{4.5pt}
    \caption{\revone{Ablation study on the DrivAerNet++ WSS task. Each row removes or modifies one component relative to the full WSS model.}}
    \label{tab:wss_ablation}
    \begin{tabular}{lcc}
        \toprule
        Variant & Rel. $L_2$ & $R^2$ \\
        \midrule
        \revone{\textbf{Full model}} & \revone{\textbf{0.226}} & \revone{\textbf{0.943}} \\
        \revone{Without shared MLP} & \revone{0.261} & \revone{0.924} \\
        \revone{Without bilinear rasterization} & \revone{0.252} & \revone{0.929} \\
        \revone{Without empty-cell smoothing} & \revone{0.252} & \revone{0.929} \\
        \revone{Without AFNO} & \revone{0.250} & \revone{0.931} \\
        \revone{Without CNN} & \revone{0.227} & \revone{0.943} \\
        \revone{Without physical PE} & \revone{0.227} & \revone{0.942} \\
        \revone{Without normal projection} & \revone{0.227} & \revone{0.942} \\
        \revone{Without curvature proxy} & \revone{0.226} & \revone{0.943} \\
        \revone{Without normal + curvature} & \revone{0.227} & \revone{0.942} \\
        \revone{Reduced backbone/query configuration} & \revone{0.247} & \revone{0.932} \\
        \bottomrule
    \end{tabular}
\end{table}

\subsection{Discussion}
\begin{figure}[!t]
    \centering
    \includegraphics[width=\linewidth]{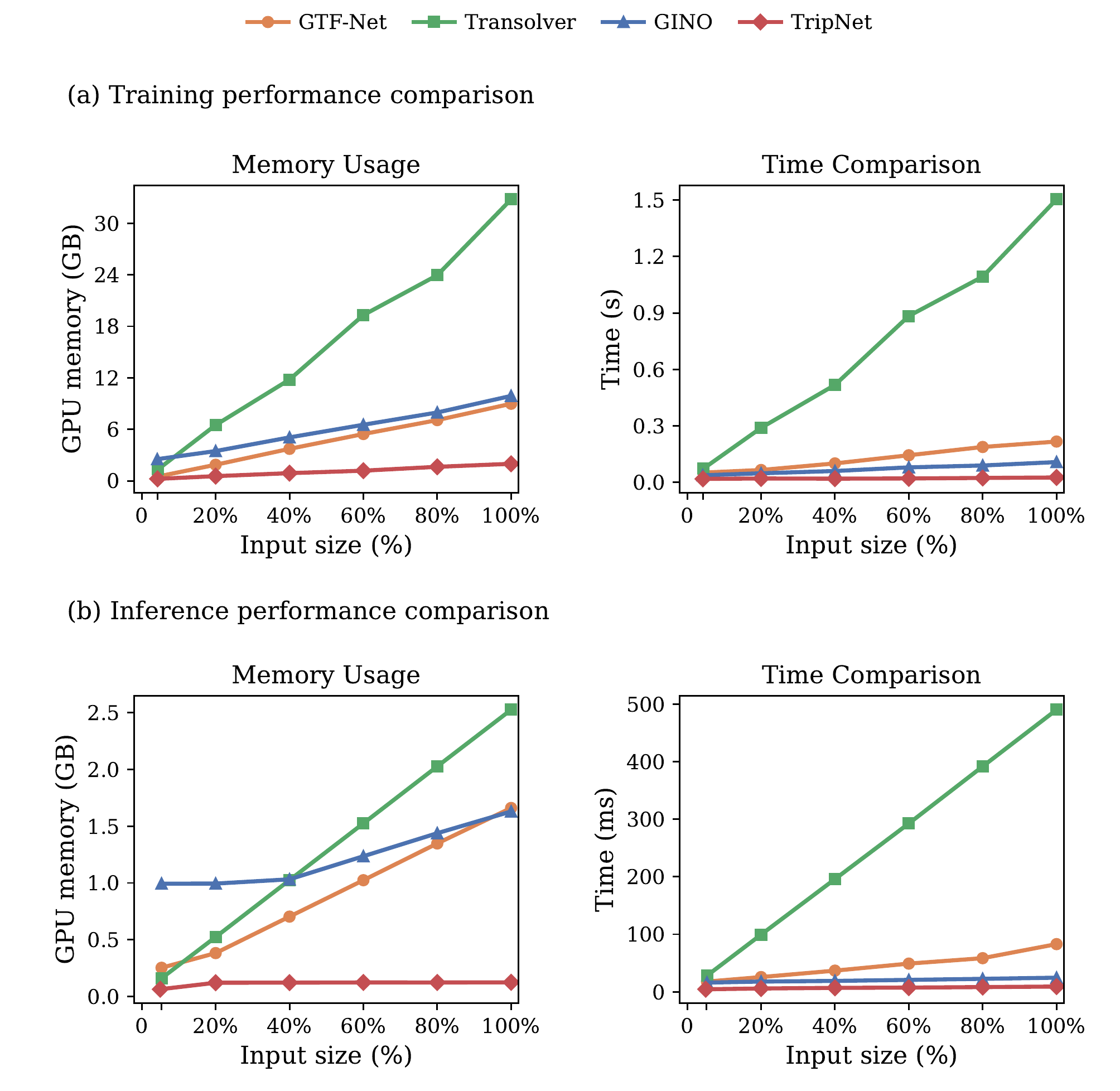}
    \caption{Computational scaling comparison between GTF-Net and baseline models on a single A100 GPU as the input surface-sample size varies from a 20{,}000-point anchor to 20\%, 40\%, 60\%, 80\%, and 100\% of the full mesh. The figure reports memory and time for training and inference; throughput curves are omitted for readability.}
    \label{fig:efficiency}
\end{figure}

The ablation results highlight two levels of complementarity in the model design. At the backbone level, removing either branch degrades all metrics, with a larger drop when the CNN branch is removed. This observation suggests that local convolutional refinement is particularly important for resolving pressure variations near geometric transitions, while the AFNO stream still contributes global spectral mixing over the vehicle surface.

At the decoder level, the three sub-components of the geometric query encoding---\allowbreak directional coordinates, normal projections, and curvature proxy---\allowbreak are not independently additive but interact in a structured way. One plausible interpretation is that the curvature proxy helps the decoder determine where normal orientation is aerodynamically informative; without this cue, the normal signal may introduce more noise than useful conditioning.

For computational costs, the proposed model is trained for 200 epochs in approximately 5.0 hours on a single A100 GPU. Fig.~\ref{fig:efficiency} compares the computational scaling of GTF-Net and baseline models as the input surface-sample size varies from a 20{,}000-point anchor to 20\%, 40\%, 60\%, 80\%, and 100\% of the full mesh. The most pronounced widening gap is observed against Transolver: at the 20k-point setting, Transolver is about 1.4 times slower per training step and 1.6 times slower at inference, whereas near full resolution it is about 6.9 times slower in training, uses 3.7 times more training memory, and is 5.9 times slower at inference. GINO shows a flatter time curve but uses comparable training memory at full resolution, while TripNet remains computationally light in this profiling but gives lower prediction accuracy in Table~\ref{tab:main_results}. These differences indicate that the efficiency advantage of GTF-Net is most evident against point-based or attention-heavy baselines whose cost grows directly with the number of surface samples, whereas triplane-style or operator baselines exhibit different accuracy--efficiency trade-offs.

\revone{The proposed GTF-Net can be used for the design-optimization workflow as a surrogate evaluator. For parametric modeling, a design vector defines a CAD surface; for free-form deformation, control-point or spline-lattice displacements define the surface. After each update, the surface is remeshed or sampled, and the normalized coordinates, reconstructed normals, and curvature proxy are recomputed before prediction. The predicted pressure and WSS fields can then be integrated with a common surface quadrature rule to obtain drag and lift coefficients, while geometric validity, smoothness, thickness, and manufacturability constraints are imposed by the optimizer. This workflow is a proposed integration route rather than an optimization experiment performed in the present study.}

Several limitations remain. First, the baseline set covers attention, operator-learning, and triplane models but is not exhaustive, partly because public DrivAerNet++ surface-field implementations remain limited. Second, the evaluation is restricted to the \revone{fixed-condition} DrivAerNet++ setting, and generalization to other vehicle datasets or flow conditions has not yet been tested. \revone{It is crucial to extend the present model to extrapolate across freestream speeds, yaw angles, ground or wheel boundary conditions, or turbulence models.}
\section{Conclusions}
We propose a geometry-aware triplane field network (GTF-Net) for high-fidelity aerodynamic surface-field prediction that refines the standard pipeline at three points: a shared MLP with smooth bilinear rasterization replaces direct coordinate projection, a dual-stream backbone combining AFNO spectral mixing and convolutional layers processes the structured triplane representation, and a geometry-aware query decoder reintroduces vehicle-aligned directional encoding, normal projections, and curvature cues at prediction time.

On the DrivAerNet++ surface-field benchmark, the method improves aggregate metrics over Transolver, GINO, and TripNet on both pressure and wall shear stress prediction, reducing relative $L_2$ error from the strongest baseline value of 0.157 to 0.145 for pressure and from 0.237 to \revone{0.225} for WSS. The model trains in approximately 5.0 hours on a single A100 GPU. Targeted ablations show that the AFNO stream, CNN branch, and geometric query encoding all contribute to the performance. In particular, removing the CNN branch gives the largest backbone-level degradation, while removing the curvature proxy alone is more damaging than disabling the entire geometric encoding branch.

Together, these results support the combination of structured triplane representations, dual-stream backbone processing, and geometry-aware query decoding for vehicle aerodynamic surface-field learning. Extending the approach to volumetric flow-field prediction, multi-condition generalization, and integration with shape-optimization or decision-oriented aerodynamic design frameworks remains a natural next step \cite{liu2026aeroagent}.

\appendix

\section{DrivAerNet++ Dataset}
DrivAerNet++ contains more than 8{,}000 high-fidelity automotive CFD cases covering three vehicle body types: Notchback, Estateback, and Fastback \cite{elrefaie2024drivaernetpp}. The official split contains 5{,}819 training samples, 1{,}148 validation samples, and 1{,}154 test samples. Each CFD case provides surface mesh geometry and corresponding surface-field quantities. In this work, the prediction targets are surface pressure and wall shear stress. The full-resolution surface mesh contains approximately $5\times10^5$ points per vehicle, while each training sample is randomly subsampled to 20{,}000 surface points.

\section{Baseline Models}
The following baseline models are adapted for surface-field regression on the DrivAerNet++ dataset and evaluated under the standardized protocol described in the main text.

\par\medskip\noindent\textbf{\textit{Transolver.}}\par\noindent
The Transolver model ingests per-point geometric coordinates and lifts them to a 256-dimensional hidden representation with a multilayer perceptron \cite{wu2024transolver}. A learnable global placeholder vector is added to the hidden representation before transformer processing. The network stacks five transformer blocks with eight attention heads. In each block, the physics-aware attention module softly assigns surface points to 32 learned slices, pools features within each slice to form slice tokens, applies self-attention among the slice tokens, and projects the attended information back to the surface points. A position-wise feed-forward network with an expansion ratio of 1.5, Layer Normalization, and residual connections is used in each block. In the final block, a Layer Normalization layer followed by a linear projection produces the target surface-field value at each point.

\par\medskip\noindent\textbf{\textit{GINO.}}\par\noindent
The GINO baseline ingests normalized surface coordinates and surface normal vectors \cite{li2023gino}. Each surface point is encoded using Fourier positional features of the coordinates concatenated with the normal vector, and a pointwise multilayer perceptron maps the resulting feature to a 64-dimensional latent representation. The point features are splatted onto a $32^3$ Cartesian latent grid by trilinear weighting. Grid coordinates and a normalized density channel are concatenated with the accumulated features, and a $1\times1\times1$ convolution lifts the grid representation. The latent grid is processed by four three-dimensional operator blocks. Each block combines spectral convolution over the retained Fourier modes with a local $1\times1\times1$ convolution, followed by Group Normalization, GELU activation, residual connections, and a pointwise convolutional multilayer perceptron. For prediction, latent features are sampled back at query surface points by trilinear interpolation, concatenated with query positional encoding and the surface normal vector, and decoded by a four-layer multilayer perceptron to produce the surface pressure or wall shear stress value.

\par\medskip\noindent\textbf{\textit{TripNet.}}\par\noindent
TripNet encodes each vehicle into a triplane representation consisting of three orthogonal feature planes \cite{chen2025tripnet}. Each plane contains 32 latent channels at a resolution of $128\times128$, and the three planes are reshaped into a 96-channel two-dimensional feature map. This feature map is processed by a U-Net-style convolutional encoder--decoder with residual skip connections, Batch Normalization, GELU activations, and a base width of 64 channels. The refined feature map is reshaped back into three triplanes. For each query point, features are sampled from the three planes by bilinear interpolation and combined into a triplane feature vector. The sampled feature is concatenated with the query coordinates and passed through a four-layer multilayer perceptron with hidden dimension 256 to predict the aerodynamic field value.

\revone{For reproducibility, the baseline optimization schedules follow the implementation settings used in the common protocol. GTF-Net and GINO use AdamW with a learning rate of $10^{-3}$, weight decay of $10^{-5}$, cosine learning-rate decay, and 200 training epochs. Transolver uses Adam with a learning rate of $10^{-3}$, no weight decay, and StepLR decay at epoch 50 with $\gamma=0.1$. TripNet uses AdamW with a learning rate of $10^{-3}$, weight decay of $10^{-5}$, cosine decay, and 200 epochs for the field predictor, while its separate triplane-fitting stage uses Adam with a learning rate of $10^{-3}$. Model selection is based on the lowest validation MSE under the common split and evaluation protocol.}

\section*{Declaration of competing interest}
The authors declare that they have no known competing financial interests or personal relationships that could have appeared to influence the work reported in this paper.

\section*{Data availability}
The DrivAerNet++ dataset used in this study is publicly available through Harvard Dataverse under the Creative Commons Attribution-NonCommercial 4.0 International License (CC BY-NC 4.0). The official train, validation, and test splits provided by DrivAerNet++ were used in this work. The processed data and evaluation artifacts generated in this study are available from the corresponding authors upon reasonable request.

\section*{Acknowledgments}
This work was supported by the National Natural Science Foundation of China (NSFC) (Grant Nos. 12588301 and 12302283); the NSFC Excellence Research Group Program for ``Multiscale Problems in Nonlinear Mechanics'' (No. 12588201); the Shenzhen Science and Technology Program (Grant Nos. SYSPG20241211173725008 and KQTD20180411143441009); and the Department of Science and Technology of Guangdong Province (Grant Nos. 2019B21203001, 2020B1212030001, and 2023B1212060001). Additional support was provided by the Innovation Capability Support Program of Shaanxi (Program No. 2023-CX-TD-30), the National Key R\&D Program of China (Grant No. 2024YFF1500600), and the Center for Computational Science and Engineering of Southern University of Science and Technology. The authors wish to extend their sincere gratitude to Hui Li from Tenfong Technology Co. Ltd., Zhou Jiang and Shi Yang from Chongqing University, and Ning Chang from Ningxia University for the insightful discussions and invaluable assistance throughout this study.

\begingroup
\footnotesize
\setlength{\bibsep}{0pt plus 0.3ex}
\hypersetup{pdfborder={0 0 0}}
\bibliographystyle{elsarticle-num-titlelink}
\bibliography{refs}
\endgroup

\end{document}